%% file: main_IEEE.tex
\documentclass[letterpaper, 10 pt, conference]{ieeeconf}
\IEEEoverridecommandlockouts
\usepackage{times}

\usepackage{multicol}
\usepackage[bookmarks=true]{hyperref}
\usepackage{color}
\usepackage[utf8]{inputenc}

\usepackage{amsthm,amsmath,amssymb}
\usepackage{graphicx}
\usepackage{import}
\usepackage{dsfont}
\usepackage{todonotes}
\usepackage{amsfonts}
\usepackage{algorithm}
\usepackage{mathtools}
\usepackage{acronym}
\usepackage{algpseudocode} 
\usepackage{tabularx}
\usepackage[most]{tcolorbox}
\usepackage{diagbox}
\usepackage[utf8]{inputenc}

\usepackage{svg}
\usepackage{cite}

\title{\LARGE \bf
An Effectiveness Study Across Baseline and Learning-based Force Estimation Methods on the da Vinci Research Kit Si System}
\author{Hao Yang, Ayberk Acar, Keshuai Xu, Anton Deguet, Peter Kazanzides, Jie Ying Wu 
\thanks{This work is supported in part by an Intuitive Surgical Technology Research Grant. Hao Yang and Jie Ying Wu are with the Department of Computer Science, Vanderbilt University, TN 37212, USA. Keshuai Xu, Anton Deguet and Peter Kazanzides are with the Department of Computer Science, Johns Hopkins University, MD 21218, USA.
All correspondence should be addressed to Hao Yang {\tt\small hao.yang@vanderbilt.edu}}
}

\begin{document}

\maketitle
\thispagestyle{empty}
\pagestyle{empty}

\begin{abstract}
Robot-assisted minimally invasive surgery, such as through the da Vinci systems, improves precision and patient outcomes. However, da Vinci systems prior to da Vinci 5, lacked direct force-sensing capabilities, forcing surgeons to operate without the haptic feedback they get through laparoscopy. Our prior work restored force sensing through machine learning-based force estimation for the da Vinci Research Kit (dVRK) Classic. This study extends our previous method to the newer dVRK system, the dVRK-Si. Additionally, we benchmark the performance of the learning-based algorithm against baseline methods (which make simplifying assumptions on the torque) to study how the two systems differ. Results show the learning-based method achieves an average root-mean-square-error (RMSE) of 5.21\%, for the dVRK-Si, which is comparable to the dVRK Classic. In both systems, the learning-based method outperforms baselines, but the difference is much larger in the dVRK-Si. Nonetheless, dVRK-Si force estimation accuracy lags behind the dVRK Classic, with RMSE 2 to 3 times higher. Further analysis reveals poor PID control in the dVRK-Si. We hypothesize that this is due to the lack of gravity compensation, as unlike the dVRK Classic, the dVRK-Si is not mechanically balanced. This study advances the understanding of learning-based force estimation and is the first work to characterize the dynamics of the new dVRK-Si system.
    
\end{abstract}

\section{Introduction}
\label{sec:introduction}
\input{section/1.Introduction}

\section{Related Works}
\label{sec:related}
\input{section/2.Related}

\section{Methods}
\label{sec:force_est}
\input{section/3.Force_est}

\section{Experimental Setup}
\label{sec:experimental_setup}
\input{section/4.Experimental_setup}

\section{Results}
\label{sec:experiment}
\input{section/5.Experiment}

\section{Discussion and Conclusion}
\label{sec:conclusion}
\input{section/6.Conclusion}

\bibliographystyle{IEEEtran}
\bibliography{references}

\end{document}

%% file: section/1.Introduction.tex
Robot-assisted minimally invasive surgery improves surgical precision and improves patient care~\cite{Reddy2023AdvancementsIR}. For example, the da Vinci$^\text{\textregistered{}}$ Surgical System (Intuitive Surgical Inc., CA) enables teleoperation, reducing surgeon fatigue, while patients benefit from less pain and faster recovery rates~\cite{DiMaio2011, Reddy2023AdvancementsIR}. However, unlike a manual laparoscopic operation where the forces are transmitted directly, the da Vinci robots prior to the da Vinci 5 lack a force-sensing mechanism. In our previous work, we developed machine learning-based torque and force estimation methods~\cite{Yilmaz2020NeuralNB, Wu2021RobotFE} for the da Vinci Research Kit (dVRK Classic), an open-source mechatronic system built on the retired first-generation da Vinci system~\cite{kazanzides-chen-etal-icra-2014}.

\begin{figure}[]
\centering
\includegraphics[width=1\columnwidth]{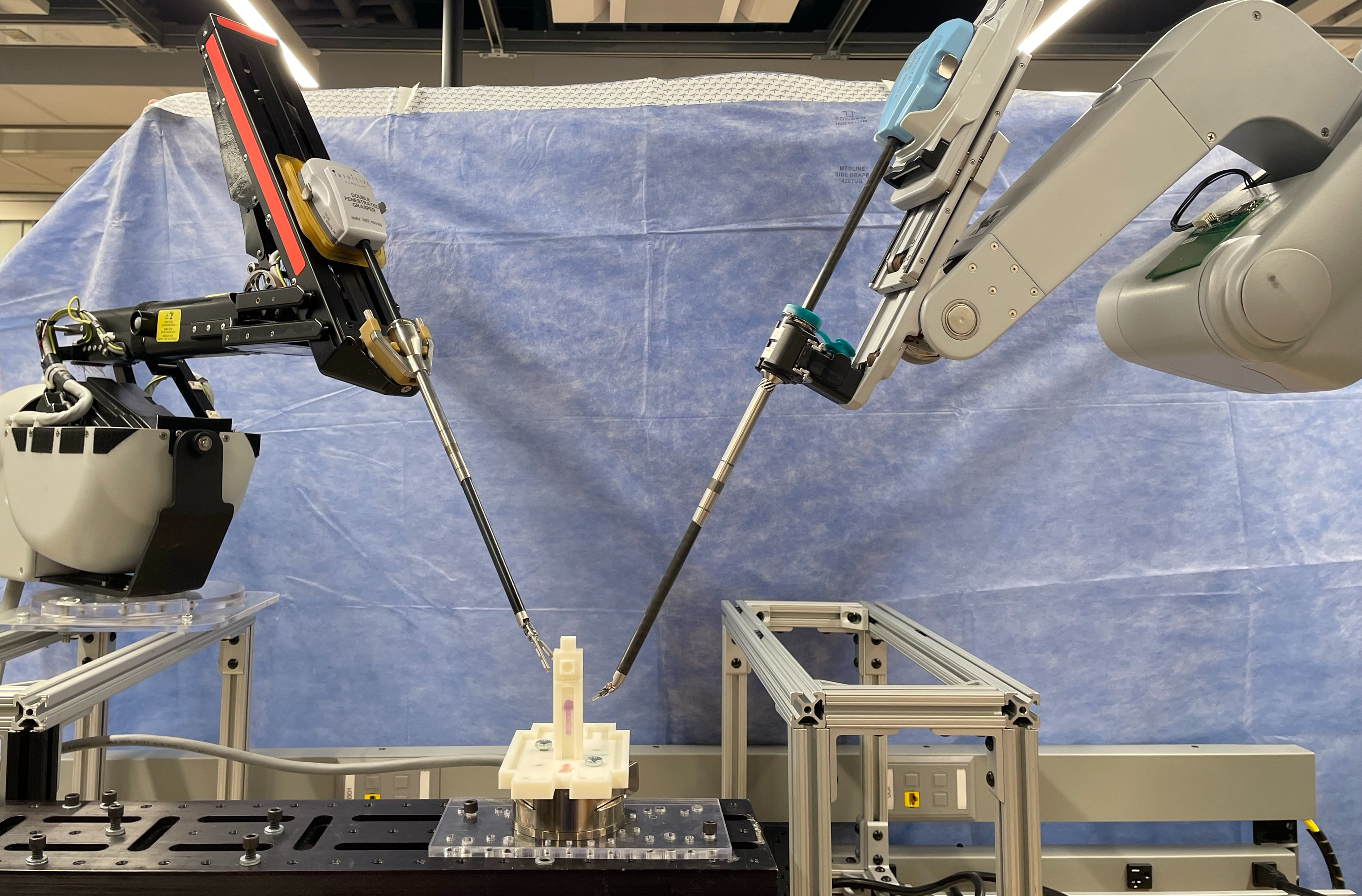}
\vspace{-4mm}
\caption{Experimental setup. From left to right are the dVRK Classic PSM arm, the F/T sensor with a 3D-printed shaft mounted on top, and the dVRK-Si PSM arm.}
\label{fig_setup}
\vspace{-2mm}
\end{figure}

In this study, we further investigate the robustness and generalization ability of our proposed learning-based method and extend it to the new generation of dVRK, the da Vinci Research Kit Si (dVRK-Si). Our experimental setup is shown in Fig.~\ref{fig_setup}. To evaluate the performance of our method, we compare the accuracy of the force estimation with several baseline methods. We assess and benchmark the effectiveness and transferability of these approaches by conducting comparative studies between the dVRK Classic Patient Side Manipulator (PSM) and dVRK-Si PSM arms. This comparative study is a step towards a force estimation pipeline for various downstream applications and tasks, such as enabling haptic feedback or estimating force applied to tissue during surgery~\cite{bergholz2023benefits}.

%% file: section/2.Related.tex
There are two main directions to approach haptic feedback in surgical robotics: embedding sensors in instruments for direct force measurement, and using model-based or learning-based methods to estimate forces indirectly. 

\subsection*{Sensor-based Methods}
Several studies have added sensors (e.g., piezoresistive sensors, load cells, or optical sensors) or modified the dVRK robot's structure to directly measure forces~\cite{King2009TactileFI,Chua2022AM3,Hosseinabadi2021UltraLF,Hosseinabadi2021UltraLowNH,Black20206DOFFS}. However, such modifications increase system complexity, affecting production and maintenance.

\subsection*{Model-based Methods}
As a cost-effective alternative, researchers have explored methods for sensorless force estimation, which can be divided into model-based and learning-based methods. The general idea is to estimate the external Cartesian force at the robot tip with the equation below:
\begin{equation} \label{eq:force}
    \hat{F}_{ext} = J^{-T}\hat{\tau}_{ext} = J^{-T} (\tau - \hat{\tau})
\end{equation}
Here, $\hat{F}_{ext}$ denotes the external Cartesian force at the tip, $\hat{\tau}$ is the estimated free-space joint torque, and $\tau$ is the measured joint torque. Sensorless force estimation methods primarily focus on estimating the free-space joint torque, $\hat{\tau}$. Zhang et al. evaluated the feasibility of using motor current for force estimation with customized 3D printed instruments~\cite{Zhao2015SensorlessFS}. Wang et al. developed a dynamic model based on convex optimization to estimate the inertial parameters of the dVRK Classic~\cite{Wang2019ACO}, later enhanced by Omer et al. with an Augmented Lagrangian Particle Swarm Optimizer~\cite{Omer2024AugmentedLagrangian}. The accuracy of these methods depends largely on the precision of robot modeling. 

\subsection*{Learning-based Methods}
Learning-based methods avoid explicit modeling of complex, nonlinear, and uncertain dynamics through learning from data. Tran et al. adopted an end-to-end approach and directly predicted $\hat{F}_{ext}$ but it generalized badly for new sensor configurations~\cite{tran2020deep}. Other works approached this by learning the free-space torque, $\hat{\tau}$, and then use Eq. \ref{eq:force} to obtain $\hat{F}_{ext}$. Shim et al. introduced a domain-unified neural network framework to enhance contact force estimation in industrial robots by addressing uncertain torque and friction~\cite{Shim2024ContactUncertain}. Yasin et al. improved joint-level ex-situ force sensing for continuum robots with advanced friction compensation and parameter updates~\cite{Yasin2020JointlevelFS}. For the dVRK, Chua et al. used a vision- and kinematics-based method to estimate tip forces of the dVRK Classic, though the required external cameras were unsuitable for surgical scenarios~\cite{Chua2021CharacterizationOR}. To improve generalizability and translation to clinical scenarios, Yilmaz et al. proposed a kinematics-only approach using a multi-layer perceptron-based neural network to estimate the free-space joint torque~\cite{Yilmaz2020NeuralNB}. Wu et al. enhanced this approach using a Long Short-Term Memory (LSTM)-based neural network for improved estimation and introduced a secondary network to compensate for the interaction forces of the patient body~\cite{Wu2021RobotFE}. Building on this series of work, we previously proposed a hybrid model- and learning-based framework for dVRK Classic force estimation, combining the generalizability of model-based methods with the flexibility of learning-based approaches~\cite{Yang2024Hybrid}. 

Previous studies show that learning-based methods achieve reasonable estimation errors. In this work, we evaluate the improvement of learning-based force estimation against baseline methods. In addition, we conduct experiments on the newly released dVRK-Si system, aiming to assess the generalizability of our approach on a distinct, uncharacterized platform. To our knowledge, this is the first work on dVRK-Si system identification.

%% file: section/3.Force_est.tex
\subsection*{Baseline Methods}
We implement three methods as baselines for comparison with the learning-based approach.
\subsubsection{Measurement Only} We ignore the free-space torque and directly use the measured torque to calculate the external Cartesian tip force. Thus, Eq.~\ref{eq:force} becomes: 
\begin{equation}
\hat{F}_{ext} = J^{-T} \tau    
\end{equation}
\subsubsection{Measurement with Bias Compensation} Assuming static forces like gravity or robot placement remain constant during an operation, we subtract the average measured torque at zero velocity from the measured torque at each timestamp. In this case, Eq.~\ref{eq:force} becomes: 
\begin{equation}
\hat{F}_{ext} = J^{-T} (\tau - {\overline \tau_{zero-velocity}})
\end{equation}
\subsubsection{Vector Search} We collect a training dataset of free-space robot movements and use it with Facebook AI Similarity Search (FAISS) as a lookup table, creating a one-to-one mapping between joint velocity-position pairs and joint torque~\cite{johnson2019billion}. For the in-contact test dataset, we estimate free-space torque offsets $\hat{\tau}$, by matching joint velocity-position pairs to their closest torque values. The equation is given by:
\begin{equation}
\hat{F}_{ext} = J^{-T} (\tau - \hat\tau_{vector-search})  
\end{equation}
\subsection*{Learning-based Method}
\begin{figure}[htbp]
    \centering
    \includegraphics[width=1\linewidth]{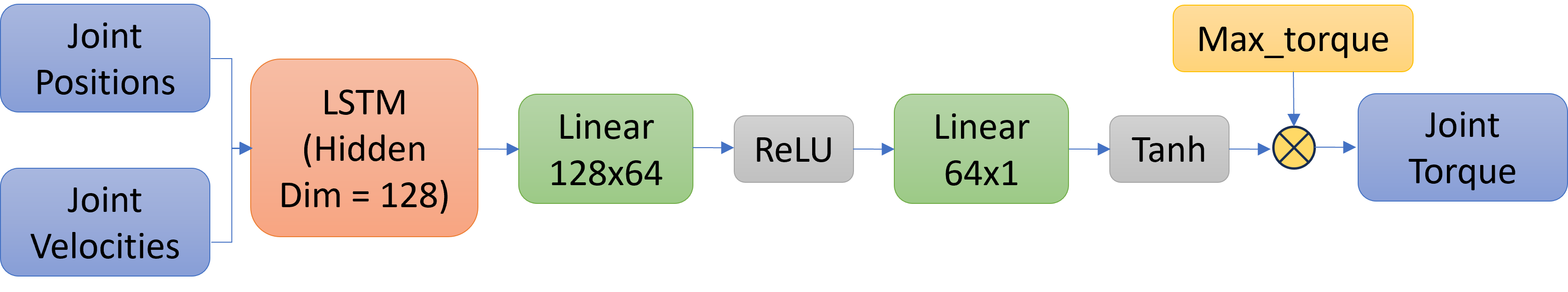}
    \caption{LSTM network architecture for one joint. Six identical networks are trained for each joint to account for different ranges of motion~\cite{Wu2021RobotFE}.}
    \label{fig:lstm}
\end{figure}

Our learning-based method follows the structure from our previous work~\cite{Wu2021RobotFE}. We train an LSTM network to take free-space joint angles and velocities as input and estimate the free-space torques for each joint of the dVRK Classic and dVRK-Si. We use L2 loss to compare the estimated and measured free-space torques. The LSTM network architecture is shown in Fig.~\ref{fig:lstm}.

We then estimate the external Cartesian force exerted on the robot instrument tip when in contact with the environment using the following equation: 
\begin{equation} \label{eq:lstm}
    \hat{F}_{ext} = J^{-T} (\tau - {\hat \tau_{LSTM}})
\end{equation}

%% file: section/4.Experimental_setup.tex
The free-space training datasets for both the dVRK Classic and dVRK-Si PSM are each collected through teleoperation for approximately 50 minutes, with the robots moving without contacting the environment. We split the datasets into 80\% for training, 10\% for validation, and 10\% for testing. We record the data at 1\,kHz, then down-sample to 200\,Hz for synchronization. We implement the network in PyTorch using the Adam optimizer (learning rate 1e-3), with a batch size of 32,764 and 1,000 epochs, applying a scheduler to decay on a plateau and cross-validation every 5 epochs~\cite{Paszke2019PyTorchAI}. The final network layer uses a Tanh activation function to regularize the output to~$\pm$1. To scale the output torque to its original range, we multiply the regularized output by a factor max\_torque, which represents the maximum torque measured in our dataset for each joint. The training takes 3-6 minutes on a desktop with an AMD Ryzen 5900x CPU and Nvidia A4500 GPU.

To test our network, we collect a separate 2.5-minute-long tip-contact test dataset, by teleoperating the robot instrument tip to contact a testing platform. The platform comprises a 3D-printed shaft mounted on an ATI Gamma F/T sensor, which provides the ground truth Cartesian forces. We teleoperate the dVRK PSM to push and pull the 3D-printed shaft from the top and sides, as shown in Fig.~\ref{fig_setup}.

%% file: section/5.Experiment.tex
\subsection*{Pre- and Post-Filtered Results}
\begin{figure}[htbp]
    \centering
    \setlength{\abovecaptionskip}{0.cm}
    \includegraphics[width=0.95\linewidth]{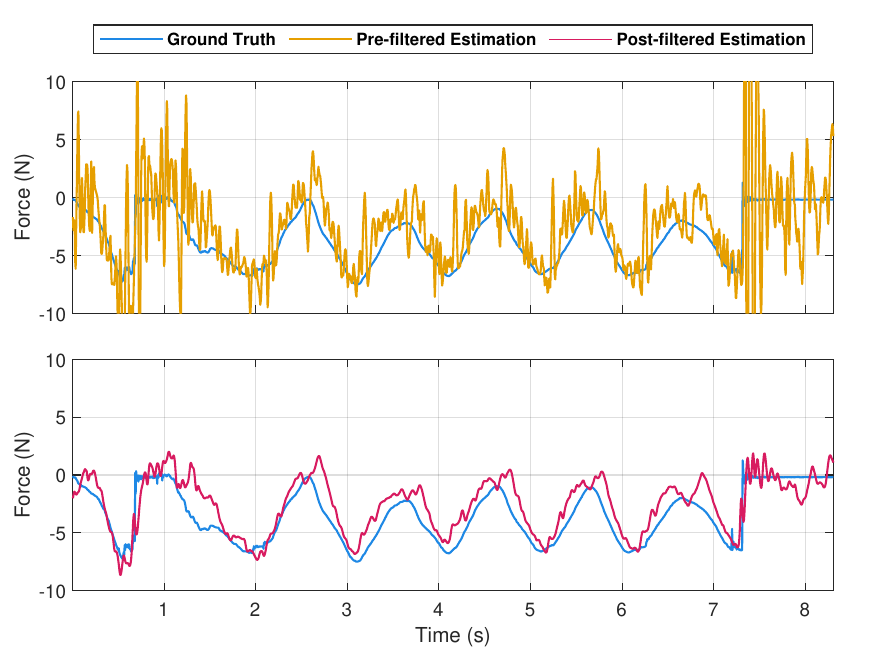}
    \caption{Comparison between pre- and post-filtered force estimation results on the x-axis. The blue curves represent the ground truth from the ATI F/T sensor. The yellow and red curves represent the estimated results.} 
    \label{fig:compare-filt}
\end{figure}

We observe that force estimation experiments reveal significantly higher residual noise in the dVRK-Si PSM compared to the dVRK Classic. To improve the force estimation accuracy, we apply a moving average filter for post-processing. The simple moving average (SMA) computes the unweighted running mean over the last $k$ entries of a data set with $n$ entries~\cite{smith1997dsp}, where the data points are $p_{1}\ldots p_{n}$. The formula for the SMA is:
\begin{align}
    \textbf{SMA}_k = \frac{1}{k} \sum_{\scriptscriptstyle \scalebox{0.6}{i=n-(k-1)}}^n p_i
\end{align}
We experimentally set $k$ to 30. We demonstrate the effectiveness of this filter by comparing the pre- and post-filtering Root Mean Square Error (RMSE) in Table~\ref{tab:filt}. The RMSE of the estimated force on the dVRK-Si PSM decreases by 24\%, 21\%, and 10\% along each axis, at the cost of an added latency of about 75\,ms. We also plot a snapshot of the pre- and post-filtered estimation curves alongside the ground truth in Fig.~\ref{fig:compare-filt} to illustrate the effect of this filtering. We show that the filtering significantly reduces the noise with minimal added latency.

\begin{table}[htbp]
\centering
\caption{RMSE of dVRK-Si Tool Tip Force Estimation}
\label{tab:filt}
\begin{tabular}{cccc}
\hline
\multicolumn{1}{|c|}{} & \multicolumn{1}{c|}{$Fx(N)$} & \multicolumn{1}{c|}{$Fy(N)$} & \multicolumn{1}{c|}{$Fz(N)$} \\ \hline\hline
\multicolumn{1}{|c|}{Pre-filtered} & \multicolumn{1}{c|}{3.27} & \multicolumn{1}{c|}{3.02} & \multicolumn{1}{c|}{1.76}  \\ \hline
\multicolumn{1}{|c|}{Post-filtered} & \multicolumn{1}{c|}{2.47} & \multicolumn{1}{c|}{2.41} & \multicolumn{1}{c|}{1.60}  \\ \hline
\end{tabular}
\end{table}

\subsection*{Force Estimation Performance Benchmark}
\begin{figure}[htbp]
    \centering
    \setlength{\abovecaptionskip}{0.cm}
    \includegraphics[width=0.95\linewidth]{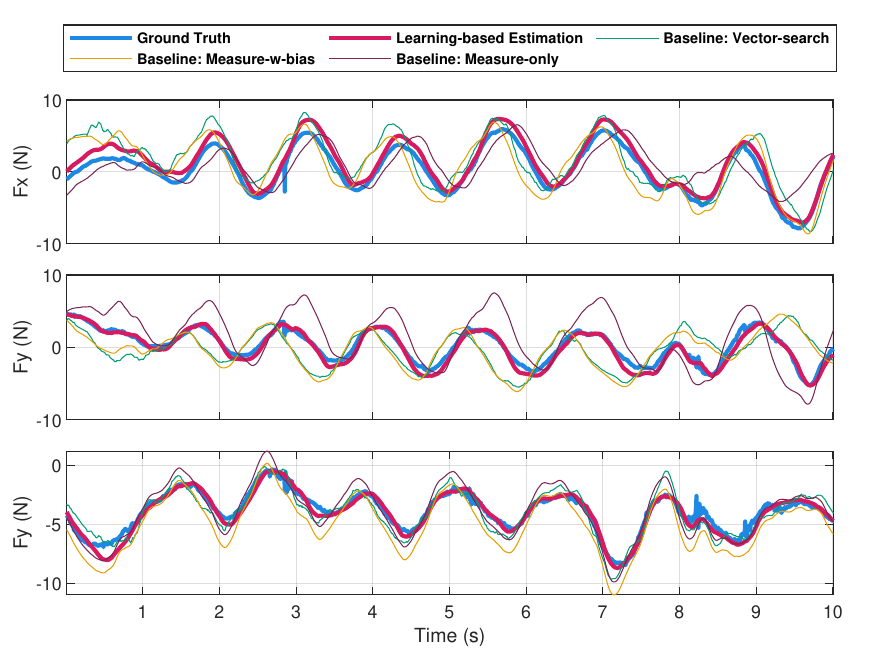}

    \includegraphics[width=0.95\linewidth]{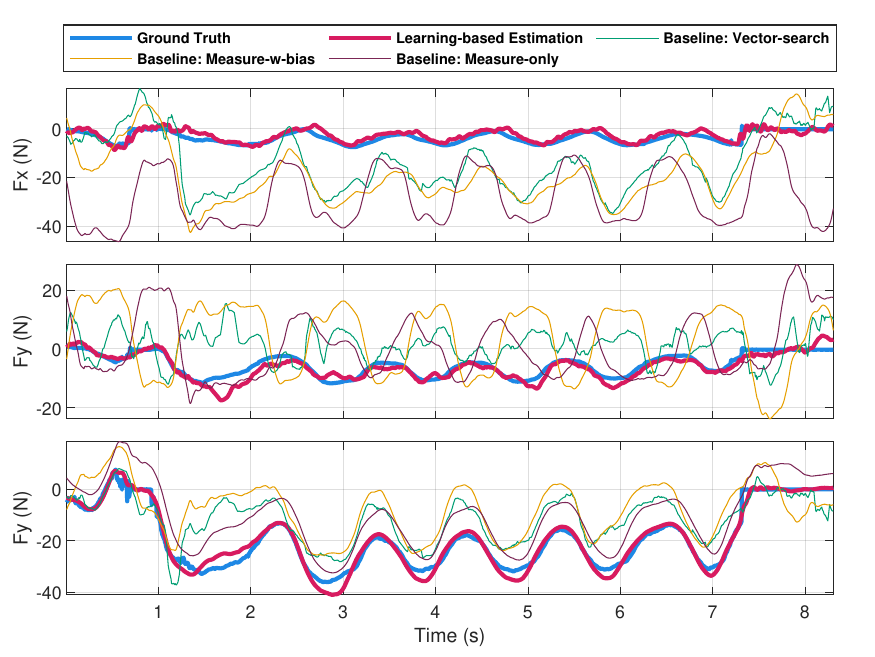}
    \caption{The baseline and learning-based method force estimation results on dVRK Classic (top) and dVRK-Si (bottom), alongside ground truth force measurements. The baseline estimations perform much worse on dVRK-Si than on dVRK Classic, as they all deviate considerably from the ground truth curve. For the X-axis, baseline estimation errors on dVRK-Si can reach 30-40\,N.}
    \label{fig:multimethods_cls_si}
\end{figure}

Table~\ref{tab:RMSE} presents the force estimation results for dVRK Classic (Cls) and dVRK-Si. Snapshots of the estimations are shown in Fig.~\ref{fig:multimethods_cls_si}, comparing learning-based and baseline methods against ground truth forces for both systems. We conclude that the learning-based method provides a comparable prediction accuracy over the two systems, as the ratio of average RMSE over the average range of force is 3.07\% for the dVRK Classic, and 5.21\% for the dVRK-Si. In addition, according to our results in Table~\ref{tab:RMSE}, we observe that the vector-search method achieves the highest accuracy among the baselines, while the learning-based method outperforms all of them.
\begin{table}[htbp]
\centering
\setlength{\tabcolsep}{4pt}
\caption{Force Estimation Benchmark: Estimation RMSE (N) and Range of Contact Force Value (N)}
\label{tab:RMSE}
\begin{tabular}{|c|c|c|c|c|c|c|}
\hline
 & \begin{tabular}[c]{@{}c@{}}Measure\\ Only\\ RMSE\end{tabular} & \begin{tabular}[c]{@{}c@{}}Measure\\ w. Bias\\ RMSE\end{tabular} & \begin{tabular}[c]{@{}c@{}}Vector\\ Search\\ RMSE\end{tabular} & \begin{tabular}[c]{@{}c@{}}NN-\\ based\\ RMSE\end{tabular} & \begin{tabular}[c]{@{}c@{}}Range\\ of\\ Force\end{tabular} & \begin{tabular}[c]{@{}c@{}} \%\\ of\\ Range\end{tabular}  \\ \hline
Fx Cls & 3.11 & 1.42 & 1.29 & \textbf{1.08} & 38.09 & 2.83\\ \hline
Fy Cls & 3.93 & 1.57 & 1.43 & \textbf{0.90} & 35.44 & 2.54\\ \hline
Fz Cls & 1.25 & 1.52 & 1.49 & \textbf{0.89} & 20.22 & 4.40\\ \hline
Ave. Cls & 2.76 & 1.50 & 1.40 & \textbf{0.96} & 31.25 & 3.07\\ \hline
Fx Si & 13.73 & 11.88 & 9.68 & \textbf{2.47} & 36.33 & 6.80 \\ \hline
Fy Si & 24.52 & 11.20 & 9.67 & \textbf{2.41} & 32.39 & 7.44\\ \hline
Fz Si & 15.56 & 3.0 & 2.94 & \textbf{1.60} & 55.69 & 2.87\\ \hline
Ave. Si & 17.94 & 8.69 & 7.43 & \textbf{2.16} & 41.47 & 5.21\\ \hline
\end{tabular}
\end{table}

\subsection*{dVRK-Si Estimation Error Analysis}

\begin{figure}[htbp]
    \centering
    \setlength{\abovecaptionskip}{0.cm}
    \includegraphics[width=0.96\linewidth]{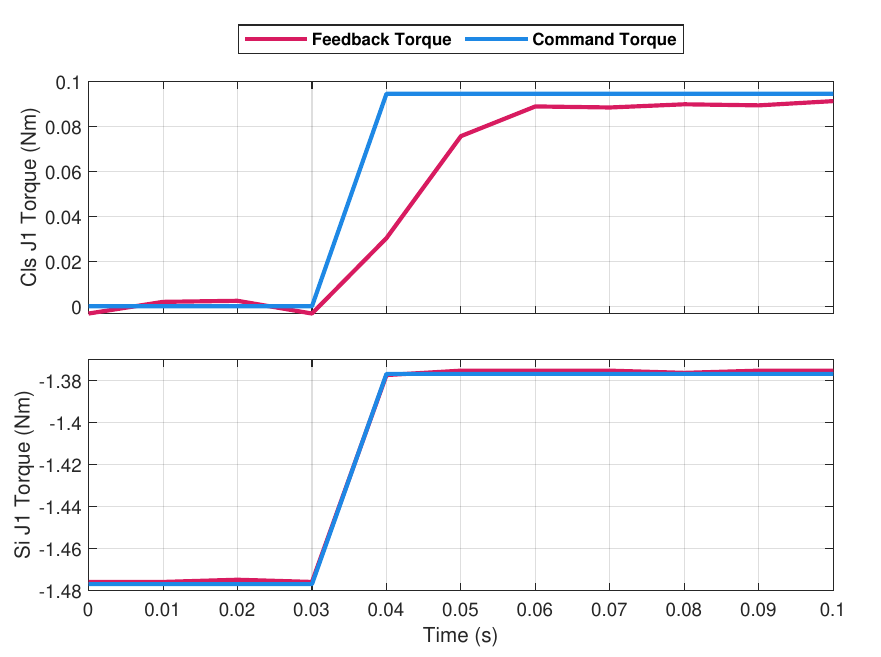}
    \caption{Comparison of transient joint torque response between the dVRK Classic and dVRK-Si. Both arms are set to home position and given a 0.1\,Nm desired torque command at 30\,ms. Frequency was set to 100\,Hz. The dVRK Classic takes 30\,ms to reach steady state, while the dVRK-Si achieves steady state within 10\,ms, matching the command timing without delay. This shows the torque response of dVRK-Si outperforms dVRK Classic. Note that dVRK Classic is self-balanced and requires no torque at the initial position, while dVRK-Si needs about -1.47\,Nm to compensate for gravity.}
    \label{fig:transient}
\end{figure}

\begin{figure}[!h]
    \centering
    \setlength{\abovecaptionskip}{0.cm}
    \includegraphics[width=0.96\linewidth]{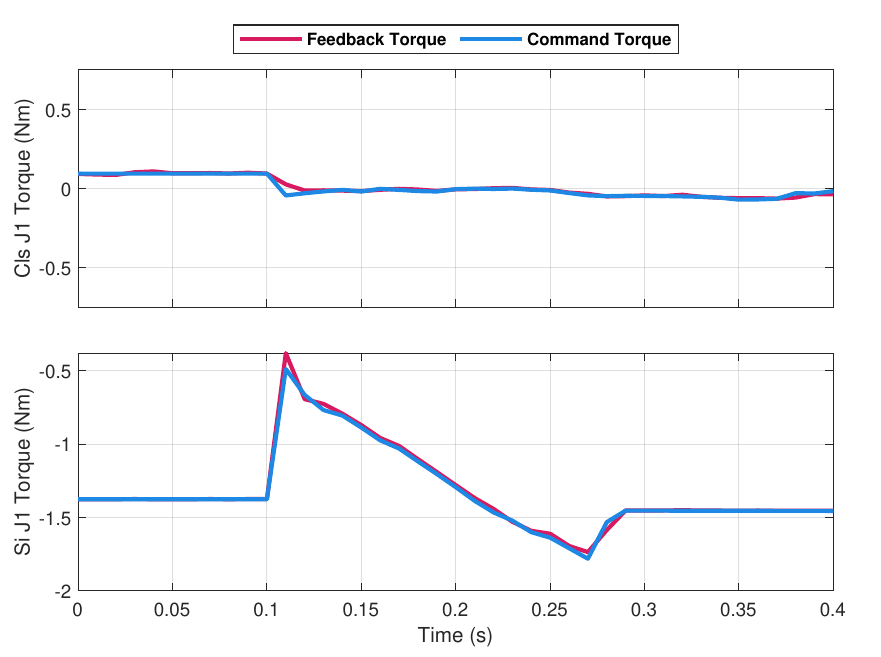}
    \caption{The PID position controller performance differs between the dVRK Classic and dVRK-Si. With data recorded at 100\,Hz, both arms receive a homing command to return the first joint to zero (home) position at 0.1\,s. The dVRK Classic, being self-balanced, requires minimal torque at home. In contrast, the dVRK-Si has stronger joints and no mechanical counterbalance, resulting in a torque of about -1.47\,Nm at the initial equilibrium state. The dVRK-Si’s higher PID gains to counter gravity and friction cause aggressive behavior, leading to overshoot of around 1.5\,Nm.}
    \label{fig:pid_position}
\end{figure}

We observe that the dVRK-Si has a large residual noise and requires filtering. We propose two hypotheses for why the dVRK-Si is noisier than the dVRK Classic: one is that the dVRK-Si uses pulse width modulation (PWM), which causes larger ripples compared to the linear amplifier of the dVRK Classic.
Another is that dVRK-Si lacks gravity compensation, leading to poor proportional–integral–derivative (PID) position controller performance. To identify the root cause of the large residual noise, we conduct joint torque transient response and PID position controller experiments.
The results are shown in Fig.~\ref{fig:transient} and Fig.~\ref{fig:pid_position}, respectively.
The dVRK-Si shows negligible delay in transient response but performs poorly in position control.


\begin{figure}[!h]
    \centering
    \setlength{\abovecaptionskip}{0.cm}
    \includegraphics[width=0.95\linewidth]{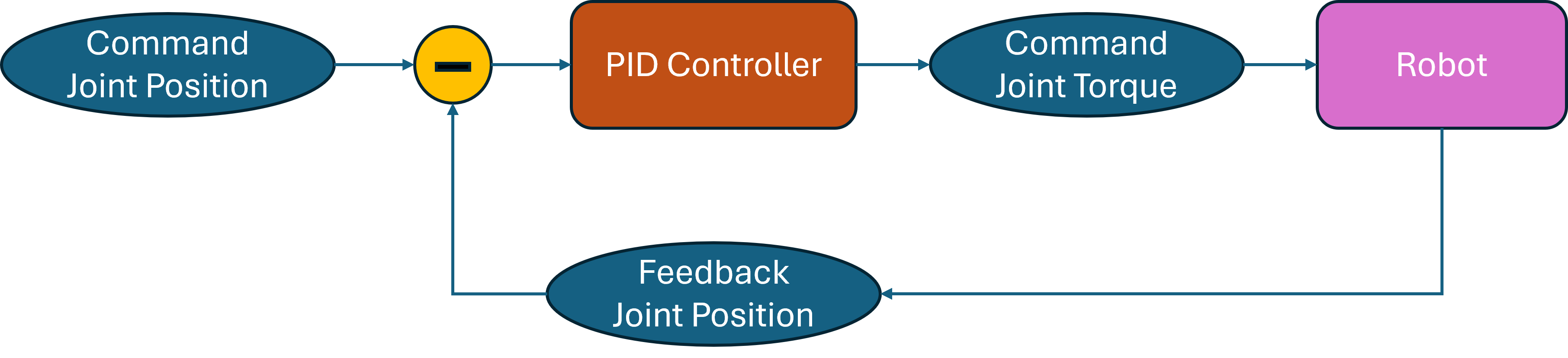}
    \caption{PID Controller structure in the dVRK software.}      
    \label{fig:pid}
\end{figure}

\begin{table}[!h]
\centering
\caption{PID Gains: dVRK Classic (Cls) PSM vs. dVRK-Si PSM}
\label{tab:pid}
\begin{tabular}{cccc}
\hline
\multicolumn{1}{|c|}{} & \multicolumn{1}{c|}{Cls/Si Joint 1} & \multicolumn{1}{c|}{Cls/Si Joint 2} & \multicolumn{1}{c|}{Cls/Si joint 3} \\ \hline\hline
\multicolumn{1}{|c|}{P Gain} & \multicolumn{1}{c|}{120/600} & \multicolumn{1}{c|}{120/600} & \multicolumn{1}{c|}{6000/6000}  \\ \hline
\multicolumn{1}{|c|}{D Gain} & \multicolumn{1}{c|}{5/30} & \multicolumn{1}{c|}{5/30} & \multicolumn{1}{c|}{200/200}  \\ \hline
\end{tabular}
\end{table}
Near the home position, the PID position controller struggles to stabilize, with an overshoot of about 1.5\,Nm and a settling time of about 0.2\,s.
This is likely due to the dVRK-Si's strong, high-gear ratio joints, whereas the dVRK Classic has lower-gear ratio joints. The dVRK Classic uses counterbalance, while the dVRK-Si does not. The dVRK-Si's design requires more torque to overcome gravity, making the PID position controller stabilization harder without gravity compensation. Note that the dVRK-Si has much larger PID gains than the dVRK Classic, as shown in Table~\ref{tab:pid}, while their controller structure is the same, as shown in Fig.~\ref{fig:pid}~\cite{kazanzides-chen-etal-icra-2014}. Poor PID control increases the residual error of the torque profile and degrades the accuracy of the force estimation of the dVRK-Si, as the quality of the training dataset relies on PID-based teleoperation and joint torque readings.

%% file: section/6.Conclusion.tex
In this study, we extend our previously proposed learning-based force estimation method to characterize a new robot, the dVRK-Si. By showing an average RMSE of 5.21\% for a range of forces that is comparable to all of our previous work on the dVRK Classic, we demonstrate that the learning-based method is effective and generalizable to the dVRK-Si system. We further validate the effectiveness of this method by comparing it to three baseline methods. We observe that the learning-based method outperforms all the baseline methods. For the dVRK-Si PSM, the RMSE of the learning-based method is 2 to 4 times smaller than the baselines. However, the dVRK-Si force estimation is less accurate than dVRK Classic, with an RMSE 2 to 3 times higher. To identify the root cause, we conduct joint torque transient response and PID position controller experiments, identifying poor PID control — likely due to the lack of gravity compensation — as the main cause of the large estimation residual error in the dVRK-Si.

